\documentclass[conference]{IEEEtran}
\IEEEoverridecommandlockouts
\usepackage{cite}
\usepackage{amsmath,amssymb,amsfonts}
\usepackage{graphicx}
\usepackage{textcomp}
\usepackage{xcolor}

\usepackage{times}
\usepackage{soul}
\usepackage{url}
\usepackage[hidelinks]{hyperref}
\usepackage[utf8]{inputenc}
\usepackage[small]{caption}
\usepackage[switch]{lineno}

\usepackage{graphicx}%
\usepackage{multirow}%
\usepackage{amsmath,amssymb,amsfonts}%
\usepackage{amsthm}%
\usepackage{mathrsfs}%
\usepackage{xcolor}%
\usepackage{textcomp}%
\usepackage{manyfoot}%
\usepackage{booktabs}%
\usepackage{algorithm}%
\usepackage{algpseudocode}%
\usepackage{listings}%
\usepackage{pifont}
\usepackage{enumitem}
\usepackage{tabularx}
\usepackage{lipsum}
\usepackage{tabularray}
\usepackage{subfig}
\usepackage[capitalise]{cleveref}
\newcommand{\cmark}{\ding{51}}%
\newcommand{\xmark}{\ding{55}}%

\def\BibTeX{{\rm B\kern-.05em{\sc i\kern-.025em b}\kern-.08em
    T\kern-.1667em\lower.7ex\hbox{E}\kern-.125emX}}
\begin{document}
\title{ShapeFormer: Shape Prior Visible-to-Amodal Transformer-based Amodal Instance Segmentation\\
}

\author{Minh Tran$^{1}$, Winston Bounsavy$^{1}$, Khoa Vo$^{1}$, Anh Nguyen$^{2}$, Tri Nguyen$^{3}$, Ngan Le$^{1}$ 
\thanks{$^{1}$AICV Lab, Department of EECS, University of Arkansas, USA.
        {\tt\small {minht}@uark.edu}}%
\thanks{$2$ Department of CS, University of Liverpool, UK.}
\thanks{$3$ Cruise LLC, USA.}
}


\maketitle

\begin{abstract}
Amodal Instance Segmentation (AIS) presents a challenging task as it involves predicting both visible and occluded parts of objects within images. Existing AIS methods rely on a bidirectional approach, encompassing both the transition from amodal features to visible features (amodal-to-visible) and from visible features to amodal features (visible-to-amodal). 
Our observation shows that the utilization of amodal features through the amodal-to-visible can confuse the visible features due to the extra information of occluded/hidden segments not presented in visible display. Consequently, this compromised quality of visible features during the subsequent visible-to-amodal transition. 
To tackle this issue, we introduce ShapeFormer, a decoupled Transformer-based model with a visible-to-amodal transition. It facilitates the explicit  relationship between output segmentations and avoids the need for amodal-to-visible transitions. ShapeFormer comprises three key modules: (i) Visible-Occluding Mask Head for predicting visible segmentation with occlusion awareness, (ii) Shape-Prior Amodal Mask Head for predicting amodal and occluded masks, and (iii) Category-Specific Shape Prior Retriever aims to provide shape prior knowledge. Comprehensive experiments and extensive ablation studies across various AIS benchmarks demonstrate the effectiveness of our ShapeFormer. The code is available at: \url{https://github.com/UARK-AICV/ShapeFormer}

\end{abstract}

\begin{IEEEkeywords}
Amodal Instance Segmentation, Shape Prior, Transformer
\end{IEEEkeywords}

\section{Introduction}
\begin{figure}[!t]
    \centering
    \includegraphics[width=\columnwidth]{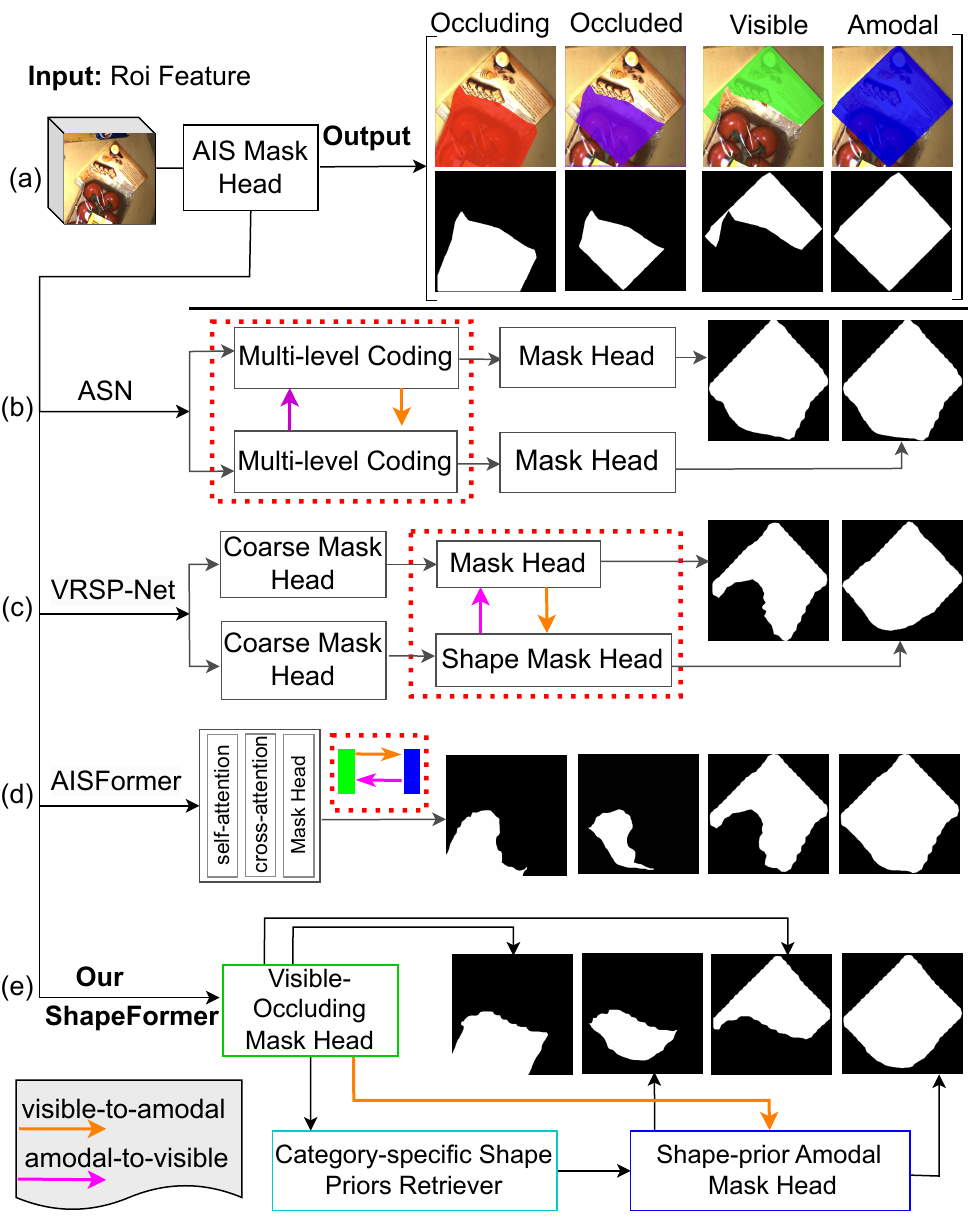}
    \caption{Comparison between our ShapeFormer and existing SOTA approaches. (a) AIS setting, which takes a RoI feature as input and returns four masks including occluding, occluded, visible and amodal. (b) ASN \cite{qi2019amodal}: bidirectional learning at multi-level coding via feature concatenation. (c) VRSP-Net \cite{xiao2021amodal}: bidirectional learning at mask head via feature concatenation. (d) AISFormer \cite{tran2022aisformer}: bidirectional learning at embeddings via self-attention. (e) Our \textbf{ShapeFormer} omits the amodal-to-visible transition, leverages the precise visible feature and shape prior knowledge to predict amodal mask.} 
    \vspace{-2em}
    \label{fig:teaser}
\end{figure}

Human perception grants us the remarkable ability to comprehend objects in their entirety i.e., an ability known as amodal perception \cite{kellman1991theory}. Based on such observation, pioneering work by \cite{zhu2017semantic,li2016amodal} introduced the concept of \textit{amodal instance segmentation (AIS)}, which focuses on determining the complete shape of an object, including its visible and occluded regions. 
Indeed, AIS holds significant potential in various applications, including robot manipulation \cite{back2022unseen} and autonomous driving \cite{qi2019amodal}. As depicted in \cref{fig:teaser}(a), AIS aims to produce the segment mask of the visible part of chocolate box (\textit{visible}), the entire chocolate box (\textit{amodal}), even when a part of it was occluded by a bag of tomatoes. The segment of this bag of tomatoes is considered as an\textit{occluding} mask. The mask joined between the amodal and occluding region is considered as an \textit{occluded} mask. In AIS, visible and amodal masks are obligated while occluding and occluded masks are supplemental outputs. 

The literature has witnessed the emergence of numerous approaches \cite{li2016amodal,follmann2019learning,qi2019amodal,mohan2022amodal,tran2022aisformer,jang2020learning,xiao2020amodal} which address the AIS challenge across various benchmarks \cite{follmann2019learning,qi2019amodal,zhu2017semantic}. 
These methods typically utilize a bidirectional approach for feature learning, involving transitions between amodal and visible features in both directions – from amodal to visible (amodal-to-visible) and from visible to amodal (visible-to-amodal). As illustrated in \cref{fig:teaser}, recent existing techniques endeavor to capture the interplay between visible-to-amodal and amodal-to-visible relationships through mechanisms like feature concatenation as in ASN \cite{qi2019amodal} and VRSP-Net \cite{xiao2021amodal}, or self-attention in AISFormer \cite{tran2022aisformer}.
However, our examination of these approaches indicates that their predictions of visible masks fall short. This inadequacy is evident in \cref{fig:teaser} (b), (c), and (d), where their visible masks result results exhibit notable deficiencies. 
We hypothesize that the amodal-to-visible relation introduces confusion into the visible masks prediction because unlike visible masks, amodal masks include regions that are not presented in the image display \cite{duncan1984selective}.
Consequently, the utilization of amodal features to enhance visible features could potentially compromise the precision of visible predictions. Moreover, when the visible mask itself is inadequate, the potential of visible features to enhance amodal mask predictions remains unrealized.

To tackle the aforementioned challenge, we introduce ShapeFormer, a novel approach that focuses exclusively on the visible-to-amodal transition, departing from the bidirectional-transition approach used in existing methods. Recent research \cite{duncan1984selective, yao2022self, xiao2021amodal, jang2020learning} underscores the efficacy of incorporating shape prior information during this transition. Building on this insight, we propose integrating shape prior knowledge into ShapeFormer. Traditional shape prior AIS methods typically employ vanilla or variational autoencoders to acquire shape priors, followed by refining coarse amodal masks. However, they often overlook the importance of object categories in shape retrieval, which can lead to overfitting the shape prior model to the training dataset. In contrast, our ShapeFormer employs a category-specific vector quantized variational autoencoder to retrieve shape priors based on the visible mask and the corresponding object category \textit{id}. Furthermore, recent research AISFormer \cite{tran2022aisformer} highlights the superior effectiveness of transformer-based architectures over CNN-based ones in modeling relationships among AIS output masks. Therefore, ShapeFormer is defined as a transformer-based architecture, aligning with these advancements in modeling techniques.

In particular, \cref{fig:teaser} (e) provides an overview of our proposed ShapeFormer, which consists of three key modules: (i) Visible-Occluding (Vis-Occ) Mask Head: This module predicts the visible segmentation mask and its category \textit{id} while acknowledging occluding segmentation. (ii) Category-Specific Shape Prior Retriever (Cat-SP Retriever): Utilizing category-specific vector quantized variational autoencoder, coupled with data augmentation, to retrieve shape priors based on the visible mask and the corresponding category \textit{id}. (iii) Shape-prior Amodal (SPA) Mask Head: Instead of simply concatenating the retrieved shape prior with the visible feature and coarse amodal feature as done in previous approaches \cite{xiao2021amodal, jang2020learning}, we leverage the shape prior knowledge as a mask within a transformer decoder's masked attention module and presents a novel shape-prior masked attention mechanism. This integration empowers the model to focus on specific regions when predicting the amodal mask, thus enhancing its accuracy and performance.

In summary, our contributions are as follows:\\
\noindent
\textbullet{ }{}We introduce ShapeFormer, a novel AIS framework with a decoupled transformer-based architecture that focuses on the visible-to-amodal transition. ShapeFormer explicitly models the relation among output segmentations while omitting the amodal-to-visible transition to prevent deficiencies in visible segmentation observed in prior works.

\noindent
\textbullet{}{ }We develop Cat-SP Retriever, a category-specific vector quantized autoencoder that leverages visible mask information and pretrains discrete codebooks for each object category to effectively retrieve shape priors. Additionally, we enhance the performance of shape prior retrieving by incorporating occlusion data augmentation, enabling better generalization to different shapes and preventing overfitting.

\noindent
\textbullet{}{ }We introduce the shape-prior masked attention to decode the amodal segmentation using the retrieved shape prior. This attention mechanism enables the model to focus on relevant parts of objects when predicting the amodal mask.

\noindent
\textbullet{}{ }Comprehensive experiments across four AIS benchmarks shows that our ShapeFormer consistently outperforms previous state-of-the-art methods. We also conduct an analysis on the effectiveness of the visible-to-amodal modeling in ShapeFormer compared to bidirectional modeling baseline. Finally, extensive ablation studies are carried out to examine the contributions of our proposed Cat-SP Retriever and shape-prior masked attention to the new state-of-the-art performance set by our ShapeFormer.
\section{Related Work}
\noindent
Amodal instance segmentation involves predicting an object’s
shape, including both its visible and occluded parts.
Li and Malik \cite{li2016amodal} first propose a method to tackle AIS by enlarging
the modal bounding box following the direction of high heatmap values and synthetically adds occlusion. 
Subsequent to this pioneering work, numerous other methodologies have emerged in the literature.

Notably, ORCNN \cite{follmann2019learning} introduces amodal and visible instance mask heads, along with an additional mask head for occluded mask prediction. Building upon ORCNN, ASN \cite{qi2019amodal} incorporates a multi-level coding module for bidirectional modeling of visible and amodal features. BCNet \cite{ke2021deep} augments amodal mask prediction with an extra branch for occluding mask prediction within the bounding box. AISFormer \cite{tran2022aisformer} introduces a transformer-based mask head, showcasing the effectiveness of transformer modeling for generating AIS output masks. However, their model implicitly learn all the relationship between output masks in one transformer model. As we mentioned earlier, this modeling contains the bidirection relation between visible and amodal feature, making visible segmentation output defective, consequently impacting the quality of the amodal segmentation output. 

Recent studies \cite{xiao2021amodal,jang2020learning} highlight the advantages of incorporating shape priors into AIS. These methods leverage mask shapes as prior knowledge to enhance amodal mask predictions. VRSP-Net \cite{xiao2021amodal} predicts coarse amodal masks, retrieves shape priors through a plain autoencoder, and refines final amodal mask predictions. AmodalBlastomere \cite{jang2020learning} uses a similar approach with a variational autoencoder for blastomere and cell segmentation. Despite their advancements, these methods tend to neglect the significance of object categories when retrieving prior shapes. Furthermore, their training procedures often result in overfitting the shape prior model to the training dataset. Additionally, these approaches simply employ the shape prior by concatenating it with the visible features for refining amodal masks.

Our proposed method, ShapeFormer, exploits the strengths of both transformers and shape priors in AIS while addressing their inherent challenges. Specifically, our approach tackles bidirectional feature learning in previous works (e.g. AISFormer\cite{tran2022aisformer}, ASN\cite{qi2019amodal}, VRSP-Net\cite{xiao2021amodal}
) by decoupling the model's transition from visible to amodal with the inclusion of shape priors. Furthermore, we mitigate previous issues associated with shape-prior-based methods by introducing a category-specific shape prior retriever, coupled with occlusion copy-paste augmentation to reduce overfitting. Additionally, the incorporation of shape-prior masked attention enables effective utilization of shape priors within a transformer-base model to predict amodal segmentation.

\begin{table}[!h]
\centering
\vspace{-0.5em}
\caption{\textit{Network architecture comparison} between our proposed ShapeFormer and existing AIS approaches. AE and VAE denote Autoencoder and Variational Autoencoder.}
\vspace{-0.5em}
\setlength{\tabcolsep}{2pt}
    \renewcommand{\arraystretch}{.9}
\resizebox{.5\textwidth}{!}{
\begin{tabular}{l|cccc}
\toprule
   \multirow{2}{*}{\textbf{Methods}}      & \multirow{2}{*}{\textbf{Networks}}   & \textbf{visible} & \textbf{amodal-to} & \multirow{2}{*}{\textbf{Shape-prior}} \\ 
& & \textbf{-to-amodal} & \textbf{-visible} & \\ \midrule
ASN \cite{qi2019amodal}             & CNNs  &   \cmark                &      \cmark               &          \xmark   \\ \hline
AISFormer \cite{tran2022aisformer}       &   Transformer       &   \cmark                &      \cmark                 &    \xmark         \\ \hline
VRSP-Net \cite{xiao2021amodal}        &    CNNs      &   \cmark                &       \cmark         &     Vanilla AE          \\ \hline
\multirow{2}{*}{\textbf{ShapeFormer}}      &   \multirow{2}{*}{ Transformer}      &   \multirow{2}{*}{\cmark}                &      \multirow{2}{*}{\xmark}   &      Conditional  \\
& & & & Vector Quantized VAE\\ \bottomrule
\end{tabular}
}
\end{table}
\section{Proposed ShapeFormer}
We commence by providing an overview pipeline illustrating the integration of our ShapeFormer as the amodal mask head within an object detection framework. Subsequently, we introduce ShapeFormer, which incorporates a transformer-based approach for visible-to-amodal transition, along with shape prior modeling. Lastly, we outline the objective functions for optimizing the network during training.
\begin{figure}[!t]
    \centering \includegraphics[width=\columnwidth]{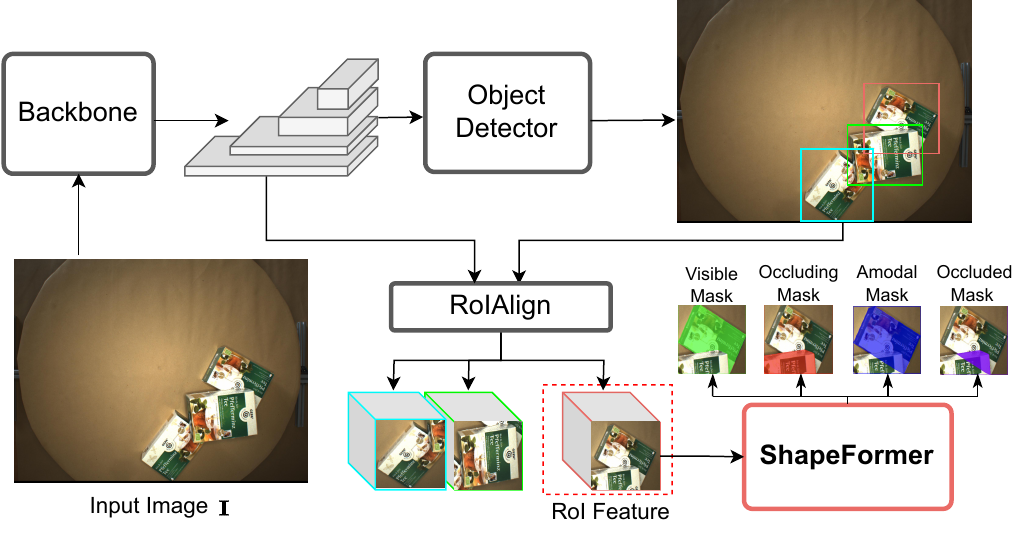}
    \caption{The overview pipeline illustrates the integration of our ShapeFormer as the amodal mask head within an object detection framework. The input image $\mathbf{I}$ goes through a backbone followed by an object detector to predict the regions of interest (RoI) and extract their corresponding feature. These RoI features are then processed through the proposed \textbf{ShapeFormer} (\cref{fig:shapeformer_maskhead}) to obtain the desired output AIS masks. }
    \vspace{-2em}
    \label{fig:overall}
\end{figure}

\subsection{Overall AIS Setup}
\cref{fig:overall} illustrates the integration of our ShapeFormer as the amodal mask head within an object detection framework. Given an input image $\mathbf{I}$, our framework follows most of previous AIS settings \cite{xiao2021amodal, follmann2019learning, ke2021deep} by utilizing a pre-trained backbone network, such as ResNet \cite{he2017mask}, RegNet \cite{schneider2017regnet} to extract spatial visual representation. An object detector such as FCOS \cite{tian2019fcos}, or Faster-RCNN\cite{he2017mask},
can be subsequently adopted to obtain $n$ regions of interest (RoI) predictions and their corresponding visual features $\{\mathbf{F}^i\}_{i=1}^n$. We also follow most of previous works \cite{xiao2021amodal, ke2021deep,tran2022aisformer}, choosing Faster R-CNN as our object detector for fair comparison. Here, each RoI is presented by its visual feature $\mathbf{F}^i \in \mathbb{R}^{C_e\times H_r\times W_r}$, where $C_e$ denotes the feature channel size and $H_r \times W_r$ represents the spatial shape of the pooling feature. 
In this context, given a RoI, our ShapeFormer takes  $\mathbf{F}^i$ as input and aims to predict the amodal mask $\mathbf{M}_a^i$, the visible mask $\mathbf{M}_v^i$, the occluding mask $\mathbf{M}_o^i$ and the occluded mask $\mathbf{M}_p^i$. 
\begin{figure*}[t]
    \centering    \includegraphics[width=1.\textwidth]{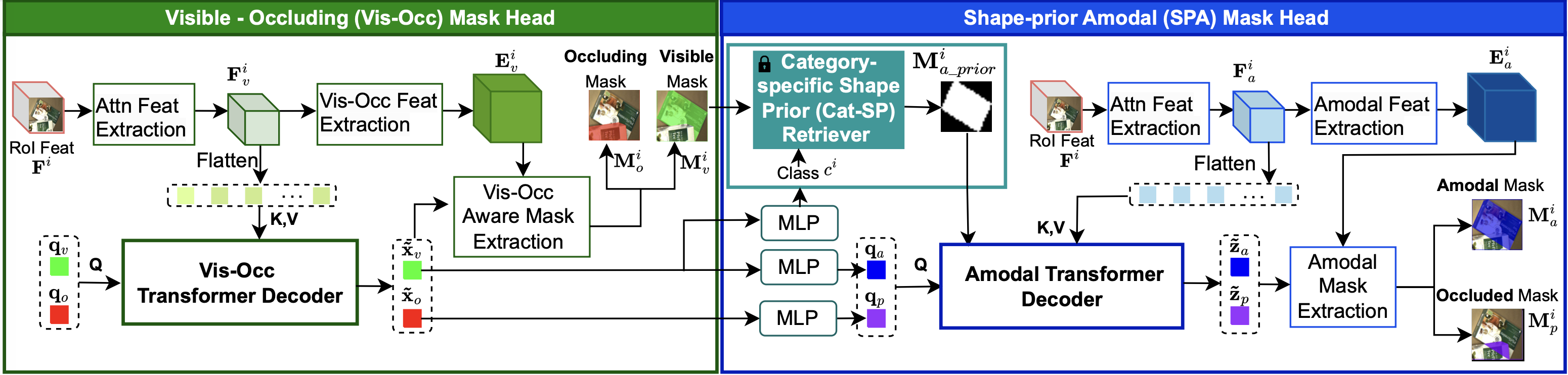}
    \caption{The pipeline of our \textbf{ShapeFormer} consisting of three main components of Visible-Occluding (Vis-Occ) Mask Head, Shape-prior Amodal (SPA) Mask Head, and Category-specific Shape Prior (Cat-SP) Retriever. Feat denotes feature.}
    \vspace{-2em}
    \label{fig:shapeformer_maskhead}
\end{figure*}

\begin{figure}[t]
    \centering    
    \includegraphics[width=\columnwidth]{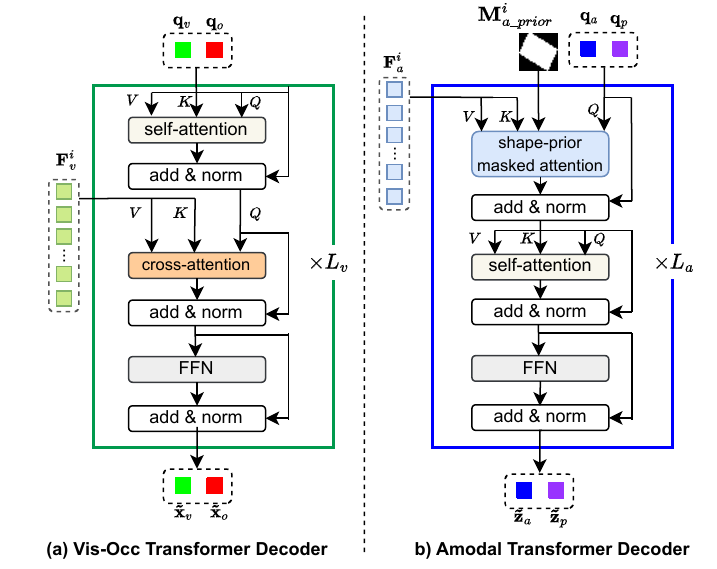}
    \caption{Detailed architecture. (a): \textbf{Vis-Occ Transformer Decoder} models the relation between visible mask and occluding mask. (b): \textbf{Amodal Transformer Decoder} with \textbf{shape-prior masked attention} models the relation between amodal mask and occluded mask. }
    \vspace{-2em}
    \label{fig:transformer_details}
\end{figure}

\subsection{ShapeFormer}
\cref{fig:shapeformer_maskhead} illustrates the key modules of our proposed ShapeFormer, which takes the RoI feature $\mathbf{F}^i$ as input. The first module, \textit{Vis-Occ Mask Head}, is designed to precisely predict the visible mask while considering occlusion (i.e. occluding mask). The second module, \textit{Cat-SP Retriever}, is responsible for retrieving a shape prior based on the visible mask and the instance's category \textit{id}. The final module, \textit{SPA Mask Head}, utilizes the shape prior and embeddings produced by the preceding modules to predict amodal mask and occluded mask.

\noindent
\subsubsection{\textbf{Vis-Occ Mask Head}}
Operating on the RoI feature $\mathbf{F}^i_v$, this module aims to make precise predictions for visible segmentation while taking occlusions into consideration. To capture the relation between the visible and the occluding masks, we introduce a transformer-based mask predictor inspired by the previous work \cite{tran2022aisformer}, which demonstrated the efficacy of relation modeling among object masks within an RoI.
In fact, we first initialize two learnable per-segment query embeddings $\mathbf{q}_v \in \mathbb{R}^{C_e}$ and $\mathbf{q}_o \in \mathbb{R}^{C_e}$ that represent the embedding of the visible segmentation and the occluding segmentation, respectively. We also extract the attention feature $\mathbf{F}^i_v$ from $\mathbf{F}^i$ by a series of three $3\times 3$ convolutional layers with a stride of $1$, followed by the extraction of Vis-Occ feature $\mathbf{E}^i_v$ by a $2\times 2$ transposed convolutional layer with a stride of $2$ plus a $1\times 1$ convolutional layer with a stride of $1$. 
Here, $\mathbf{F}^i_v$ represents key-value cross attention feature for decoding the mask embeddings whereas $\mathbf{E}^i_v$ encapsulates the semantic feature concerning whether each pixel in the RoI belongs to the visible mask of the primary object or pertains to occluding objects.
To decode the two query embeddings $\mathbf{q}_v, \mathbf{q}_o$ from the attention feature $\mathbf{F}^i_v$, we introduce the \emph{Vis-Occ Transformer Decoder} $\mathcal{D}_v$ with $L_v$ layers, as illustrated in \cref{fig:transformer_details} (a). Each layer contains one self-attention block, responsible for learning the relation between visible and occluding embeddings, followed by a cross attention block that learns the relation between the two embeddings with the attention feature $\mathbf{F}^i_v$. Formally, the decoded visible and occluding embeddings, denoted as $\mathbf{\tilde{x}}_v$ and $\mathbf{\tilde{x}}_o$, respectively,  can be computed as $[\mathbf{\tilde{x}}_v, \mathbf{\tilde{x}}_o] = \mathcal{D}_v ([\mathbf{q}_v, \mathbf{q}_o], \mathbf{F}_v^i) $. They are then correlated with everypixel embedding in $\mathbf{E}^i_v$ through a Vis-Occ Aware Mask Extraction to determine whether the pixel belongs to the visible segment or the occluding segment. The Vis-Occ Aware Mask Extraction is designed as a dot product on the feature dimension $C_e$. The output visible mask (denoted as $\mathbf{M}^i_v$), the occluding mask (denoted as $\mathbf{M}^i_o$) are formally computed as follow:
\begin{equation}
        [\mathbf{M}^i_v,\mathbf{M}^i_o]  = \text{sigmoid}([\mathbf{\tilde{x}}_v, \mathbf{\tilde{x}}_o] \otimes \mathbf{E}^i_v)
\end{equation}

\noindent
\subsubsection{\textbf{Cat-SP Retriever}}
\begin{figure}[t]
    \centering
\includegraphics[width=\columnwidth]{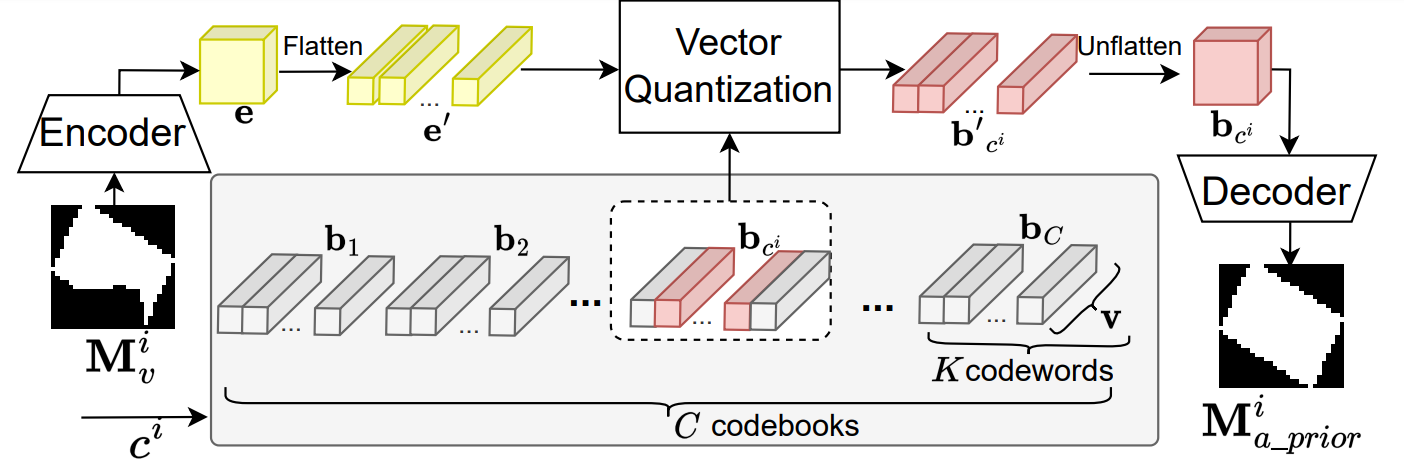}
\vspace{-1em}
    \caption{Flowchart of \textbf{Cat-SP Retriever}. Input is visible mask $\mathbf{M}_v^i$ and its class label $c^i$. Output is category-specific shape prior $\mathbf{M}^i_{a\_{prior}} = f_S (\mathbf{M}_v^i, c^i)$.} 
    \vspace{-2em}
\label{fig:shape_prior_searcher}
\end{figure}

\cref{fig:shape_prior_searcher} illustrates the overall architecture of our Cat-SP Retriever, denoted as $f_S$. It takes a visible mask $\mathbf{M}_v^i$ along with its corresponding category \textit{id} (denoted as $c^i$) as inputs. The purpose of our Cat-SP Retriever is to search for a category-specific shape prior denoted as $\mathbf{M}^i_{a\_prior}$, achieved through the operation $f_S (\mathbf{M}_v^i, c^i)$. To obtain category \textit{id} $c^i$ of $\mathbf{M}_v^i$, we employ a MLP consisting of two hidden layers followed by a softmax layer to transform the decoded visible embedding $\mathbf{\tilde{x}}_v$ into the category probability $\mathbf{p}^i \in \mathbb{R}^{C}$, where $C$ presents the total number of categories. The category \textit{id} is then obtained by applying $\arg max$ on $\mathbf{p}^i$ as below.
\begin{equation}
         c^i = \arg max~ \mathbf{p}^i \text{, where }\mathbf{p}^i = \text{softmax} (\text{MLP} (\mathbf{\tilde{x}}_v) )  
\end{equation}
We propose the utilization of a variational autoencoder with vector quantization mechanism to conduct $f_S$.
We initialize $C$ codebooks representing $C$ categories in a dataset. A codebook $\mathbf{b}_j \in \mathbb{R}^{K\times \mathbf{v}}, j \in \{1, 2, ..., C\}$ is a collection of $K$ codewords (vectors) of size $\mathbf{v}$ representing the possible latent codes for object shape of each category.
The process begins with encoding the input visible mask $\mathbf{M}_v^i$ into a encoded feature  of $\mathbf{e} \in \mathbb{R}^{k\times k\times \mathbf{v}}$ using an encoder, structured similarly to the UNet encoder \cite{ronneberger2015u}.
The encoded feature $\mathbf{e}$ is then flatten into a list of $m = k\times k$ latent vectors of size $\mathbf{v}$. These latent vectors $\mathbf{e'} \in \mathbb{R}^{m\times \mathbf{v}}$, is subjected to the quantization step. Here, the category-specific codebook $\mathbf{b}_{c^i}$ is determined based on the predicted category $c^i$ from the previous step. Each latent vector from $\mathbf{e'}$ is compared to the codewords in $\mathbf{b}_{c^i}$. The nearest codeword, determined by cosine distance, is selected as the quantized representation for each latent vector. 
After quantization, a set of $m$ codewords, denoted as $\mathbf{b'}_{c^i} \in \mathbb{R}^{m\times K}$, is selected from $\mathbf{b}_{c^i}$ to represent the $m$ encoded latent vectors in $\mathbf{e'}$. The collection $\mathbf{b'}_{c^i}$ is then unflattened to form a spatial decoded feature $\mathbf{b}_{c^i}$, which is then passed through a Unet decoder to yield the corresponding $\mathbf{M}^i_{a\_{prior}}$.

\noindent
\subsubsection{\textbf{SPA Mask Head}}
This final module is designed to predict amodal mask using occluding embedding $\mathbf{\tilde{x}}_o$, visible embedding $\mathbf{\tilde{x}}_v$ from the Vis-Occ Mask Head and shape prior $\mathbf{M}^i_{a\_{prior}}$ from Cat-SP Retriever. 
Similar to the Vis-Occ Mask Head, we also extract
the amodal attention feature $\mathbf{F}^i_a$ from $\mathbf{F}^i$, followed by the extraction of amodal feature $\mathbf{E}^i_a$, both using the same convolutional operations as in Vis-Occ Mask Head. Note that $\mathbf{F}^i_a$ represents key-value cross attention feature for decoding the mask embeddings whereas $\mathbf{E}^i_a$ encapsulates the amodal semantic feature.
It is important to note that amodal semantic also includes the occluded information, thus we also predict the occluded mask, learning from the occluding embedding $\mathbf{\tilde{x}}_o$. This enables the model to discern which parts of the occluding object obscure the amodal portion. To accomplish this, we create learnable queries for both amodal and occluded masks, denoted as $\mathbf{q}_a$ and $\mathbf{q}_p$, respectively. Due to $\mathbf{\tilde{x}}_v$ and $\mathbf{\tilde{x}}_o$ are in the visible embedding space whereas $\mathbf{q}_a$ and $\mathbf{q}_p$ are in the amodal embedding space, we propose to use MLPs to transfer those two embedding spaces, i.e. $\mathbf{q}_a = MLP (\mathbf{\tilde{x}}_v)$ and $\mathbf{q}_p = MLP (\mathbf{\tilde{x}}_o)$. We then introduce Amodal Transformer Decoder $\mathcal{D}_a$,  which incorporates shape prior $\mathbf{M}^i_{a\_{prior}}$. This decoder is responsible for decoding the amodal embedding $\mathbf{\tilde{z}}_a$ and the occluded embedding $\mathbf{\tilde{z}}_p$, as detailed in \cref{fig:transformer_details}(b). In this process, $\mathbf{q}_a$ and $\mathbf{q}_p$ are treated as queries $\mathbf{Q}$, and the amodal attention feature $\mathbf{F}^i_a$ serves as $\mathbf{K}$ and $\mathbf{V}$. 

Differing from a conventional transformer decoder that employs the traditional cross-attention mechanism, we introduce a \emph{shape-prior masked attention} within the Amodal Transformer Decoder $\mathcal{D}_a$.
This attention mechanism takes $\mathbf{Q}$, $\mathbf{K}$, $\mathbf{V}$, and the shape prior $\mathbf{M}^i_{a\_{prior}}$ as inputs. The incorporation of shape prior $\mathbf{M}^i_{a\_{prior}}$ in this masked attention allows the model to focus on specific regions, enhancing both the efficiency and effectiveness for predicting amodal segmentation. To elaborate, at each layer $l$ of the decoder, given the intermediate output from the previous layer $l-1$, denoted as $\mathbf{Z}^l = [\mathbf{z}_a^{l-1}, \mathbf{z}_p^{l-1}]$, the output of the masked attention at layer $l$ is as below:
\begin{subequations}
\begin{align}
    & \mathbf{Z}^l = \text{softmax} (\mathcal{M} + \mathbf{Q}\mathbf{K}^T) \mathbf{V} + \mathbf{Z}^{l-1} \\
    & \mathbf{Q} = \mathbf{Z}^l \cdot  \mathbf{W}^\mathbf{Q} \text{; } \mathbf{K} = \mathbf{F}_a^i \cdot \mathbf{W}^\mathbf{K} \text{ ; } \mathbf{V}  = \mathbf{F}_a^i \cdot \mathbf{W}^\mathbf{V}\\
    & \mathcal{M}(x, y) = \left\{\begin{array}{ll}
  0  & \text{if~} \mathbf{M}_{a\_prior}(x,y)=1 \\
    -\infty & \text{otherwise} \end{array}\right.
\end{align}
\end{subequations}
Here, $\mathbf{W}^\mathbf{Q}$, 
$\mathbf{W}^\mathbf{K}$,
$\mathbf{W}^\mathbf{V}$ 
are learning parameters of query $\mathbf{Q}$, key $\mathbf{K}$ and value $\mathbf{V}$, respectively. Following the masked attention, the process continues with self-attention, which aims to capture the correlation between the amodal and occluded embeddings. The decoding process of $\mathcal{D}_a$ can be expressed as follow: 
\begin{equation}
    [\mathbf{\tilde{z}}_a, \mathbf{\tilde{z}}_p] = \mathcal{D}_a ([\mathbf{q}_a, \mathbf{q}_p], \mathbf{F}_a^i, \mathbf{M}_{a\_prior})
\end{equation}
where
$\mathbf{\tilde{z}}_a$ and $\mathbf{\tilde{z}}_p$ are then correlated with each pixel embedding in $\mathbf{E}^i_a$  through an Amodal Mask Extraction, which is designed as a dot product on the feature dimension $C_e$ to derive the amodal and the occluded masks. In summary, the output amodal mask (denoted as $\mathbf{M}^i_a$), the occluded mask (denoted as $\mathbf{M}^i_p$) are computed as follow:
\begin{equation}
        \mathbf{M}^i_a  = \text{sigmoid}(\mathbf{\tilde{z}}_a \otimes \mathbf{E}^i_a); \mathbf{M}^i_p  = \text{sigmoid}(\mathbf{\tilde{z}}_p \otimes \mathbf{E}^i_a)
\end{equation}
\subsection{Training Process}
\noindent
\subsubsection{\textbf{Training Cat-SP Retriever}}
To achieve representative codebooks for our Cat-SP Retriever, we employ the training process optimizing the following objective functions. 
\begin{align}
\begin{split}
    & \mathcal{L}_{csp} = \mathcal{L}_{rec} + \mathcal{L}_{vq} \\ 
    & \mathcal{L}_{rec} = MSE (\mathbf{M}^i_{a\_{prior}}, \mathbf{M}^i_{a\_gt})\\ & \mathcal{L}_{vq}  = MSE (\mathbf{e}', \mathbf{b'}_{c^i})
\end{split}
\end{align}
Here, the reconstruction loss, denoted as $\mathcal{L}_{rec}$, is calculated by computing the mean square error (MSE) between the predicted shape prior $\mathbf{M}^i_{a\_{prior}}$ and the corresponding ground truth amodal mask $\mathbf{M}^i_{a\_gt}$. 
Meanwhile, the vector quantization loss, $\mathcal{L}_{vq}$ is optimized to learn
the codewords in the selected codebook to better match the flatten encoded feature $\mathbf{e}'$. 
\begin{figure}[t]
    \centering
\includegraphics[width=1\columnwidth]{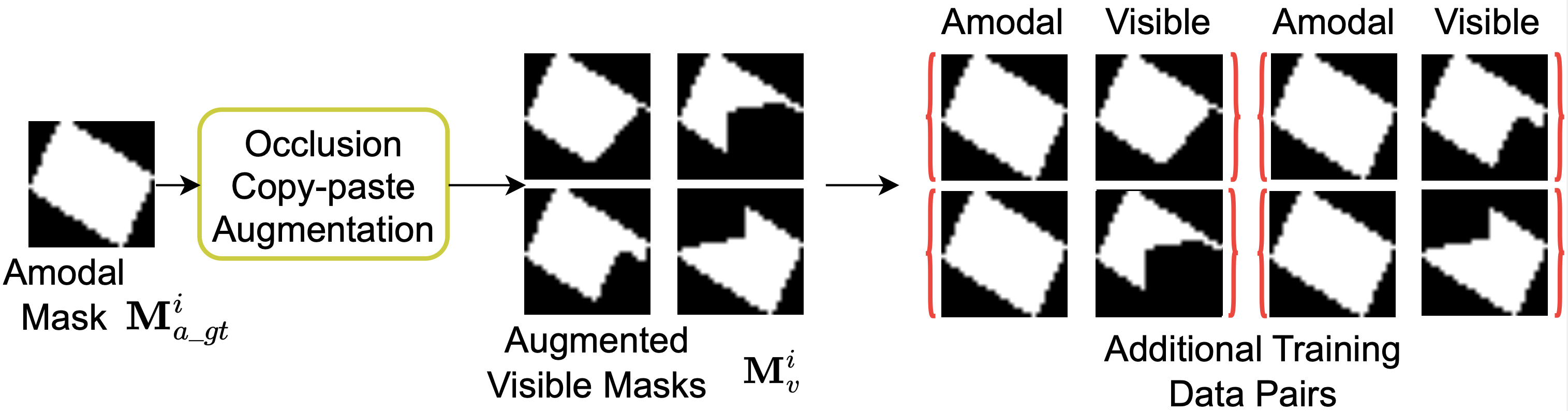}
\vspace{-1em}
    \caption{Generation of augmented visible masks ($\mathbf{M}^i_v$) from ground-truth amodal masks ($\mathbf{M}^i_{a\_gt}$) during training Cat-SP Retriever.} 
    \vspace{-2em}
\label{fig:Augmentation}
\end{figure}
Additionally, during the training of $f_S$, we generate augmented visible masks from the ground-truth amodal mask. This augmentation helps enhancing the generalization of Cat-SP Retriever by covering more occlusion scenarios that can occur during testing. Examples showcasing our augmented data can be seen in \cref{fig:Augmentation}.
Our $f_S$ is pretrained and  remains fixed during during the training of ShapeFormer, serving as a consistent source of shape prior knowledge throughout the training process.

\noindent
\subsubsection{\textbf{Training ShapeFormer}}
Our ShapeFormer is trained in an end-to-end manner concurrently with the object detection framework. 
Our training follows AIS protocols as shown in Fig.2, employing a two-stage instance segmentation process similar to Mask R-CNN. This approach enables the concurrent training of both the bounding box and amodal mask prediction heads without the need for pre-bootstrapping in object detection. 
In other words, the training procedure optimizes a multi-task loss function $\mathcal{L}$ as follow:
\begin{align}
    \mathcal{L} = \mathcal{L}_{det} + \mathcal{L}_{{cls}} + \mathcal{L}_v +\mathcal{L}_o + \mathcal{L}_a + \mathcal{L}_p
\end{align}
where $\mathcal{L}_{det}$ is object detection loss, defined similarly to that in Faster R-CNN object detection. The occluding mask loss $\mathcal{L}_o$, the visible mask loss $\mathcal{L}_v$, the amodal mask loss $\mathcal{L}_a$, occluded mask loss $\mathcal{L}_p$ and the classification loss $\mathcal{L}_{cls}$ are computed as follow:
\begin{align}
\begin{split}
    \mathcal{L}_o &= \mathcal{L}_{BCE} (\mathbf{M}^i_o, \mathbf{M}^i_{o\_gt} ),
    \mathcal{L}_v = \mathcal{L}_{BCE} (\mathbf{M}^i_v, \mathbf{M}^i_{v\_gt} ) \\
    \mathcal{L}_a &= \mathcal{L}_{BCE} (\mathbf{M}^i_a, \mathbf{M}^i_{a\_gt} ),\mathcal{L}_p = \mathcal{L}_{BCE} (\mathbf{M}^i_p, \mathbf{M}^i_{p\_gt} ) \\
    \mathcal{L}_{{cls}} &= \mathcal{L}_{CE} (\mathbf{p}^i, c^i_{gt} )
\end{split}
\end{align}
Here, $\mathbf{M}^i_{o\_gt}, \mathbf{M}^i_{v\_gt}, \mathbf{M}^i_{p\_gt} $, and $ \mathbf{M}^i_{a\_gt}$, are the ground truth of the occluding, visible, occluded and amodal masks, respectively. $c^i_{gt}$ is the ground-truth category of the RoI. $\mathcal{L}_{BCE}$ denotes the binary cross entropy loss whereas $\mathcal{L}_{CE}$ denotes the cross entropy loss.

\section{Experiments}
\subsection{Datasets, Metrics and Implementation Details}
\noindent
\textbf{Datasets:}
We benchmark our ShapeFormer on four AIS datasets, namely KINS \cite{qi2019amodal}, COCOA \cite{zhu2017semantic}, COCOA-cls \cite{follmann2019learning}, and D2SA \cite{follmann2019learning}. KINS is a large-scale traffic dataset with 95,311 training instances and 92,492 testing instances with 7 categories. COCOA is an AIS dataset that is derived from MSCOCO \cite{lin2014microsoft} with no categories, including 15,139 training instances and 8,279 testing instances. COCOA-cls is proposed to capture 80 categories object category in COCOA, however, it has much fewer annotation (6,763 training instances and 3,799 testing instances). D2SA is an AIS dataset with 60 categories of instances related to supermarket items with 13,066 training instances and 15,654 testing instances.

\begin{table}
    \caption{\textit{Performance comparison on KINS test set} with various backbones. $\dagger$ indicates our reproduced results.}
    \vspace{-0.5em}
    \centering
    \setlength{\tabcolsep}{2pt}
    \renewcommand{\arraystretch}{.9}
    \resizebox{.5\textwidth}{!}{
    \begin{tabular}{c|l|c|c|cc|cccc}
    \toprule
         \multicolumn{2}{l|}{\multirow{2}{*}{\textbf{Backbones\& Methods}}}  & \multirow{2}{*}{\textbf{Venue}}&\textbf{{Shape}} & \multicolumn{2}{c|}{\textbf{Visible}} & \multicolumn{4}{c}{\textbf{Amodal}} \\ \cmidrule{5-10}
        ~ & ~ & ~ & \textbf{{Prior}} & $AP\uparrow$ & $AR\uparrow$ & $AP\uparrow$ & $AP_{50}\uparrow$ & $AP_{75}\uparrow$ & $AR\uparrow$ \\ \midrule
        \multirow{9}{*}{\rotatebox{90}{ResNet-50}} & PCNet\cite{zhan2020self} & CVPR20   & \xmark & - & - & $29.1$ & $51.8$ & $29.6$ & $18.3$ \\ 
        ~ & ASBU\cite{nguyen2021weakly} & ICCV21  & \xmark & - & - & $29.3$ & $52.1$ & $29.7$ & $18.4$ \\ 
        ~ & Mask R-CNN\cite{ke2022mask} & ICCV17 & \xmark & $28.0$ & $19.2$ & $30.0$ & $54.5$ & $30.1$ & $19.4$ \\  
        ~ & ORCNN\cite{follmann2019learning} & WACV19  &  \xmark & $28.8$ & $20.0$ & $30.6$ & $54.2$ & $31.3$ & $19.7$ \\ 
        ~ & ASN\cite{qi2019amodal} & CVPR19  & \xmark & - & - & $32.2$ & - & - & - \\ 
        ~ & AISFormer\cite{tran2022aisformer} & BMVC22  & \xmark & ${29.7}$ & $\underline{20.0}$ & $\underline{33.8}$ & $\underline{57.8}$ & $\underline{35.3}$ & $\underline{21.1}$ \\ 
        ~ & AmodalBlastomere\cite{jang2020learning} & TMI20   & \checkmark & - & - & $30.3$ & - & - & - \\ 
        ~ & VRSP-Net\cite{xiao2021amodal}& AAAI21  & \checkmark & $\underline{29.9}$ & $19.9$ & $32.1$ & $55.4$ & $33.3$ & $20.9$ \\ \cmidrule{2-10}
        ~ & \textbf{ShapeFormer(Ours)}& -   & \checkmark & $\textbf{31.3}$ & $\textbf{21.1}$ & $\textbf{34.1}$ & $\textbf{58.6}$ & $\textbf{35.7}$ & $\textbf{22.0}$ \\ \midrule
        \multirow{5}{*}{\rotatebox{90}{ResNet-101}} & Mask R-CNN\cite{he2017mask} $\dagger$& ICCV17  & \xmark & - & - & $30.2$ & $54.3$ & $30.4$ & $19.5$ \\ 
        ~ & BCNet\cite{ke2021deep} & CVPR21  & \xmark & - & - & $28.9$ & - & - & - \\ 
        ~ & BCNet\cite{ke2021deep}  $\dagger$ & CVPR21 &  \xmark & - & - & $32.6$ & $57.2$ & $35.4$ & $21.5$ \\ 
        ~ & AISFormer\cite{tran2022aisformer}& BMVC22  & \xmark & $\underline{30.9}$ & $\underline{20.1}$ & $\underline{34.6}$ & $\underline{58.2}$ & $\underline{36.7}$ & $\underline{21.9}$ \\ \cmidrule{2-10}
        ~ & \textbf{ShapeFormer(Ours)}& -  &  \checkmark & $\textbf{32.6}$ & $\textbf{22.3}$ & $\textbf{35.2}$ & $\textbf{59.3}$ & $\textbf{37.2}$ & $\textbf{23.0}$ \\ \midrule 
        \multirow{3}{*}{\rotatebox{90}{RegNet}} & ASPNet\cite{mohan2022amodal}& CVPR22  & \xmark & - & - & $\underline{35.6}$ & - & - & - \\ 
        ~ & AISFormer\cite{tran2022aisformer} & BMVC22  &  \xmark & $\underline{31.9}$ & $\underline{21.1}$ & $\underline{35.6}$ & $\underline{59.9}$ & $\underline{37.0}$ & $\underline{22.5}$ \\ \cmidrule{2-10}
        ~ & \textbf{ShapeFormer(Ours)} & - & \checkmark & $\textbf{33.7}$ & $\textbf{22.8}$ & $\textbf{36.1}$ & $\textbf{59.9}$ & $\textbf{38.7}$ & $\textbf{23.0}$ \\ 
        \bottomrule
    \end{tabular}
    \vspace{-1.8em}
    }
    \label{tab:kins}
\end{table}

\begin{table}
    \caption{\textit{Performance comparison on COCOA test set} with various backbones. $\dagger$ indicates our reproduced results.}
    \vspace{-0.5em}
    \centering
    \setlength{\tabcolsep}{2pt}
    \renewcommand{\arraystretch}{.85}
    \resizebox{.5\textwidth}{!}{
    \begin{tabular}{c|l|c|c|cc|cccc}
    \toprule
        \multicolumn{2}{l|}{\multirow{2}{*}{\textbf{Backbones\& Methods}}} & \multirow{2}{*}{\textbf{Venue}} & \multirow{2}{*}{\textbf{\shortstack{Shape \\ Prior}}} & \multicolumn{2}{c|}{\textbf{Visible}} & \multicolumn{4}{c}{\textbf{Amodal}} \\ \cmidrule{5-10}
 ~ & ~ & ~ & ~ & $AP\uparrow$ & $AR\uparrow$ & $AP\uparrow$ & $AP_{50}\uparrow$ & $AP_{75}\uparrow$ & $AR\uparrow$ \\ \midrule
\multirow{5}{*}{\rotatebox{90}{ResNet-50\space}}  & PCNet\cite{zhan2020self} & CVPR20 & \xmark & - & - & $22.6$ & $46.8$ & $19.7$ & $6.3$ \\ 
        ~ & ASBU\cite{nguyen2021weakly}  & ICCV21 & \xmark & - & - & $23.8$ & $47.9$ & $21.2$ & $6.4$ \\ 
        ~ & ORCNN\cite{follmann2019learning}  $\dagger$ & WACV19 & \xmark & $\underline{32.9}$ & $\underline{9.3}$ & $34.8$ & $\textbf{62.9}$ & $35.1$ & $9.6$ \\ 
        ~ & AISFormer\cite{tran2022aisformer}  $\dagger$ & BMVC22 & \xmark & $32.3$ & $9.2$ & $\underline{35.6}$ & $62.5$ & $\underline{36.3}$ & $\underline{9.7}$ \\ \cmidrule{2-10}
        ~ & \textbf{ShapeFormer(Ours)} & - & \checkmark & $\textbf{33.2}$ & $\textbf{9.4}$ & $\textbf{35.7}$ & $\underline{62.7}$ & $\textbf{36.6}$ & $\textbf{9.9}$ \\  \midrule
        \multirow{5}{*}{\rotatebox{90}{ResNet-101}}  & Amodal MRCNN\cite{follmann2019learning} & WACV19 & \xmark & $29.4$ & - & $35.6$ & - & - & - \\ 
        ~ & ORCNN\cite{follmann2019learning}  & WACV19 & \xmark & $30.0$ & - & $30.1$ & - & - & - \\ 
        ~ & ORCNN\cite{follmann2019learning}  $\dagger$ & WACV19 & \xmark & $\underline{34.0}$ & $\underline{9.3}$ & $36.5$ & $64.5$ & $37.2$ & $10.0$ \\ 
        ~ & AISFormer\cite{tran2022aisformer} $\dagger$ & BMVC22 & \xmark & $33.7$ & $\underline{9.3}$ & $\underline{37.3}$ & $\underline{64.7}$ & $\underline{38.6}$ & $\underline{10.3}$ \\ \cmidrule{2-10}
        ~ & \textbf{ShapeFormer(Ours)} & - & \checkmark & $\textbf{34.7}$ & $\textbf{9.7}$ & $\textbf{37.8}$ & $\textbf{65.1}$ & $\textbf{39.4}$ & $\textbf{10.4}$ \\ \bottomrule
    \end{tabular}
    \vspace{-0.5em}
    }
    \label{tab:cocoa-test}
\end{table}
\noindent
\textbf{Metrics:} 
Following existing AIS approaches \cite{xiao2021amodal,follmann2019learning, tran2022aisformer}, we adopt mean average precision (AP) and mean average recall (AR). To evaluate our Cat-SP retriever, we adopt Intersection over Union (IoU) metric between retrieved shape priors and ground-truth shape.\\
\noindent
\textbf{Implementation Details:}
We implement our ShapeFormer based on Detectron2 \cite{wu2019detectron2}. 
For the KINS dataset, we use an SGD optimizer \cite{ruder2016overview} with a learning rate of 0.0025 and a batch size of 1 on 48000 iterations. For D2SA datasets, we also train with an SGD optimizer but with a learning rate of 0.005 and a batch size of 2 on 70000 iterations. For COCOA and COCOA-cls,  we train on 10000 iterations with the learning rate of 0.0005 and a batch size of 2.
All experiments have been conducted using an Intel(R) Core(TM) i9-10980XE 3.00GHz CPU  and a Quadro RTX 8000 GPU.

\subsection{Performance Comparison}
\begin{table}[!t]
    \caption{\textit{Performance comparison on D2SA test set} with ResNet-50 as backbone. $\dagger$ indicates our reproduced results.}
    \vspace{-0.5em}
    \centering
    \setlength{\tabcolsep}{4pt}
    \renewcommand{\arraystretch}{.9}
    \resizebox{0.5\textwidth}{!}{
    \begin{tabular}{l|c|c|cc|cccc}
    \toprule
         \multirow{2}{*}{\textbf{Methods}} & \multirow{2}{*}{\textbf{Venue}} & \multirow{2}{*}{\textbf{\shortstack{Shape \\ Prior}}} & \multicolumn{2}{c|}{\textbf{Visible}} & \multicolumn{4}{c}{\textbf{Amodal}} \\ \cmidrule{4-9}
  ~ & ~ & ~ & $AP\uparrow$ & $AR\uparrow$ & $AP\uparrow$ & $AP_{50}\uparrow$ & $AP_{75}\uparrow$ & $AR\uparrow$ \\ \midrule
                Mask R-CNN\cite{he2017mask}  & ICCV17 & \xmark & $68.98$ & $70.11$ & $63.57$ & $83.85$ & $68.02$ & $65.18$ \\ 
        ORCNN\cite{follmann2019learning}  & WACV19 & \xmark & $69.67$ & $70.46$ & $64.22$ & $83.55$ & $69.12$ & $65.25$ \\ 
        ASN\cite{qi2019amodal}  $\dagger$ & CVPR19 & \xmark & - & - & $63.94$ & $84.35$ & $69.57$ & $65.20$ \\ 
        BCNet\cite{ke2021deep}  $\dagger$ & CVPR21 & \xmark & - & - & $65.97$ & $84.23$ & $72.74$ & $66.90$ \\ 
        AISFormer\cite{tran2022aisformer}  & BMVC22 & \xmark & $71.60$ & $71.59$ & $67.22$ & $84.05$ & $72.87$ & $68.13$ \\ 
        VRSP-Net\cite{xiao2021amodal} & AAAI21 & \checkmark & $\underline{72.28}$ & $\underline{71.85}$ & $\underline{70.27}$ & $\underline{85.11}$ & $\underline{75.81}$ & $\underline{69.17}$ \\ \midrule
        \textbf{ShapeFormer(Ours)} & - & \checkmark & $\textbf{73.78}$ & $\textbf{73.05}$ & $\textbf{71.03}$ & $\textbf{86.05}$ & $\textbf{76.13}$ & $\textbf{69.31}$ \\ \bottomrule
    \end{tabular}
    \vspace{-0.5em}
    }
    \label{tab:d2sa}
\end{table}

\subsubsection{\textbf{Quantitative Results and Comparison}}
In the following tables, on each backbone, the best scores are in \textbf{bold} and the second best scores are in \underline{underlines}.
\begin{table}[!t]
    \caption{\textit{Performance comparison on COCOA-cls test set}, ResNet-50 as backbone. $\dagger$ indicates our reproduced results.}
    \vspace{-0.5em}
    \centering
    \setlength{\tabcolsep}{4pt}
    \renewcommand{\arraystretch}{.9}
    \resizebox{0.5\textwidth}{!}{
    \begin{tabular}{l|c|c|cc|cccc}
    \toprule
         \multirow{2}{*}{\textbf{Methods}} & \multirow{2}{*}{\textbf{Venue}} & \multirow{2}{*}{\textbf{\shortstack{Shape \\ Prior}}} & \multicolumn{2}{c|}{\textbf{Visible}} & \multicolumn{4}{c}{\textbf{Amodal}} \\ \cmidrule{4-9}
  ~ & ~ & ~ & $AP\uparrow$ & $AR\uparrow$ & $AP\uparrow$ & $AP_{50}\uparrow$ & $AP_{75}\uparrow$ & $AR\uparrow$ \\ \midrule
                Mask R-CNN\cite{he2017mask} & ICCV17 & \xmark & $30.10$ & $31.52$ & $33.67$ & $56.50$ & $35.78$ & $34.18$ \\ 
        ORCNN\cite{follmann2019learning} & WACV19 & \xmark & $30.80$ & $32.23$ & $28.03$ & $53.68$ & $25.36$ & $29.83$ \\ 
        ASN\cite{qi2019amodal}  $\dagger$ & CVPR19 & \xmark & - & - & $35.33$ & $58.82$ & $37.10$ & $35.50$ \\ 
        BCNet\cite{ke2021deep} $\dagger$ & CVPR21 & \xmark & - & - & $35.14$ & $\textbf{58.84}$ & $36.65$ & $35.80$ \\ 
        AISFormer\cite{tran2022aisformer} & BMVC22 & \xmark & $34.00$ & $\textbf{36.44}$ & $\underline{35.77}$ & $57.95$ & $38.23$ & $36.71$ \\        
        VRSP-Net\cite{xiao2021amodal}  & AAAI21 & \checkmark & $\underline{34.58}$ & $\underline{36.42}$ & $35.41$ & $56.03$ & $\underline{38.67}$ & $\underline{37.11}$ \\ \midrule 
        \textbf{ShapeFormer(Ours)} & - & \checkmark & \textbf{35.01} & \underline{36.42} & \textbf{35.83} & \underline{58.82} & \textbf{38.85} & \textbf{37.13} \\ \bottomrule
    \end{tabular}
    \vspace{-0.5em}
    }
    \label{tab:cocoa-cls}
\end{table}

\noindent
\textbf{KINS.} \cref{tab:kins} presents a comparison between ShapeFormer and SOTA AIS methods on the KINS dataset. ShapeFormer demonstrates consistent improvements across various backbone architectures, including ResNet-50~\cite{he2016deep}, ResNet-101~\cite{he2016deep} and RegNet~\cite{schneider2017regnet}. Specifically, when compared to methods utilizing ResNet-50 as the backbone, our method outperforms both SOTA shape-based methods 
(e.g., and VRSP-Net~\cite{xiao2021amodal} by 1.4 visible AP and 2.0 amodal AP) and non-shape-based methods (e.g., AISFormer~\cite{tran2022aisformer} by 1.6 visible AP and 0.3 amodal AP), respectively. 
When ResNet-101 is utilized as the backbone, our method achieves a larger margins of improvement over AISFormer, outperforming it by 1.7 in terms of visible AP and 0.6 in terms of amodal AP.
Furthermore, compared to APSNet \cite{mohan2022amodal} and AISFormer~\cite{tran2022aisformer} on the RegNet backbone, our approach achieves SOTA performance by surpassing them in visible AP by 1.8 and amodal AP by 0.5.\\
\noindent
\textbf{COCOA.} We also conduct experiments on COCOA test set in \cref{tab:cocoa-test}. Our ShapeFormer achieves best performance on most metrics across backbones. ShapeFormer surpasses the SOTA AISFormer  by 0.1 in amodal AP and 0.9 in visible AP when evaluated on ResNet 50. Additionally, it achieves a 0.5 improvement in amodal AP and a perfect 1.0 in visible AP when assessed on ResNet 101.

\noindent
\textbf{D2SA.} \cref{tab:d2sa} further validates our approach on D2SA dataset. We achieve best results across all metrics. Specifically, we gains 1.5 on visible AP and 0.76 on amodal AP in comparison with the second best method, i.e. VRSP-Net.\\
\noindent
\textbf{COCOA-cls.} \cref{tab:cocoa-cls} shows our results on COCOA-cls dataset.
Our ShapeFormer outperform across other methods on visible and amodal AP metrics and show competitive results on AR metrics.

In summary, our experimental results across datasets demonstrate that our approach, which incorporates visible-to-amodal modeling with shape prior, delivers comprehensive and competitive performance in both visible and amodal AP.

\subsubsection{\textbf{Qualitative results and comparison}} 
\begin{figure}[!t]
    \centering
    \includegraphics[width=0.9\linewidth]{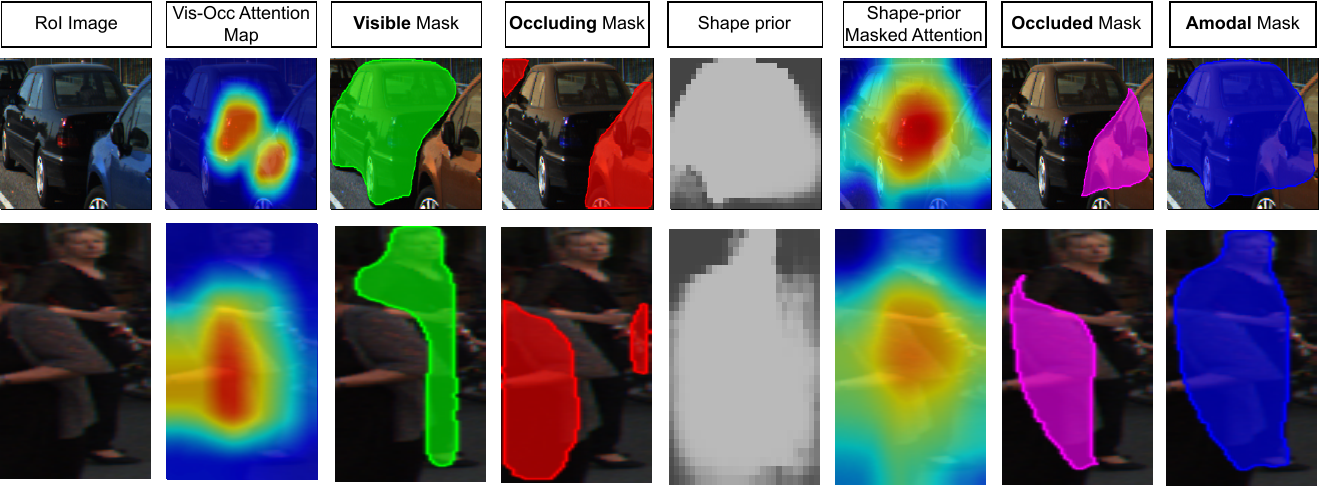}
    \caption{Qualitative results of ShapeFormer. Left to right: Input RoI, Vis-Occ attention map, Visible masks, Occluding masks, Shape priors, Shape-prior masked attention, Amodal masks, and Occluded masks.}
    \vspace{-1em}
    \label{fig:quali_results}
\end{figure}

\begin{figure}[t]
    \centering    
    \includegraphics[width=0.9\linewidth]{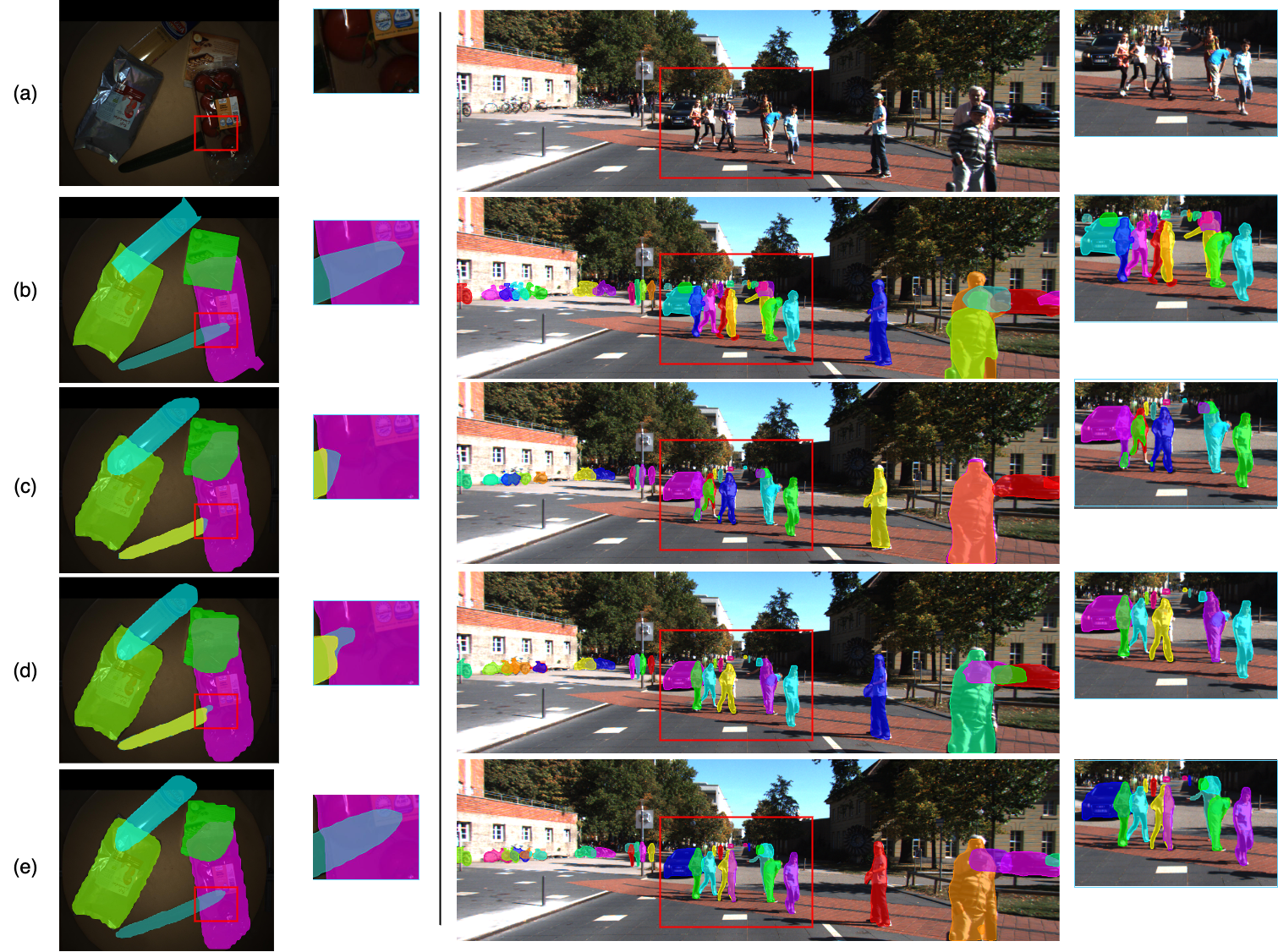}
    \caption{Quantitative comparison between our ShapeFormer and state-of-the-art methods (e.g. AISFormer~\cite{tran2022aisformer}, VRPS-Net~\cite{xiao2021amodal}) on {amodal segmentation} results. From top to bottom, (a) Image, (b) Ground truth, (c) VRSP-Net's predictions, (d) AISFormer's predictions, (e) Our ShapeFormer's prediction.  Images are sampled from D2SA (left) and KINS (right) test sets. Best view in zoom and color.}
    \vspace{-1.8em}
    \label{fig:amodal_compare}
\end{figure}

\cref{fig:quali_results} illustrates the qualitative output of ShapeFormer. To explain where the network learns, we also visibly include attention maps corresponding to Vis-Occ attention map in Vis-Occ Mask Head module and Shape-prior Masked Attention in SPA Mask Head module.
This visualization offers a comprehensive overview of both the output masks and the corresponding attention maps generated during the prediction process of our ShapeFormer model. The results are arranged from left to right, encompassing: input RoIs, Vis-Occ Attention Maps, Visible Masks, Occluding Masks, Shape priors, Shape-prior Masked Attention, Amodal masks, and Occluded masks.
\cref{fig:amodal_compare} shows qualitative comparison between our ShapeFormer and existing SOTA methods (e.g. AISFormer\cite{tran2022aisformer}, VRPS-Net\cite{xiao2021amodal}). Images are sampled from D2SA and KINS test sets. As can be seen, our ShapeFormer accurately extracts the amodal mask of the occluded object (i.e. the cucumber) (left) and efficiently handles the dense group of pedestrians (right).

\subsection{Ablation Experiments \& Analysis}
\begin{figure}[!t]
    \centering
    \includegraphics[width=.45\textwidth]{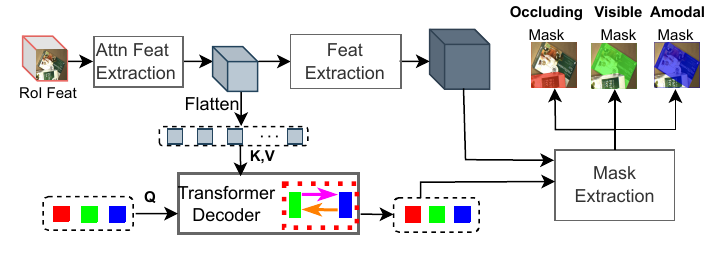}
    \vspace{-1em}
    \caption{Baseline with a bidirectional-transition. The baseline implementation shares the same design as the Vis-Occ Mask Head but includes the integration bidirectional relationships.} 
    \label{fig:ablation_baseline}
\end{figure}
\begin{table}[!t]
    \caption{The effectiveness of \textit{visible-to-amodal modeling} without shape prior (w/o SP) compared to bidirectional modeling baseline (Fig. \ref{fig:ablation_baseline}).}
    \vspace{-0.5em}
    \centering
    \setlength{\tabcolsep}{1pt}
    \renewcommand{\arraystretch}{.9}
    \resizebox{.50\textwidth}{!}{
    \begin{tabular}{l|cc|cc|cc|cc}
    \toprule
        \multirow{3}{*}{\textbf{Models}} &  \multicolumn{4}{c|}{\textbf{D2SA}} & \multicolumn{4}{c}{\textbf{KINS}} \\ \cline{2-9}
        ~ & \multicolumn{2}{c|}{\textbf{Visible}} & \multicolumn{2}{c|}{\textbf{Amodal}} &  \multicolumn{2}{c|}{\textbf{Visible}}& \multicolumn{2}{c}{\textbf{Amodal}}  \\ \cline{2-9}
         ~&  $AP\uparrow$ & $AR\uparrow$ & $AP\uparrow$ & $AR\uparrow$ & $AP\uparrow$ & $AR\uparrow$ & $AP\uparrow$ & $AR\uparrow$ \\ \midrule
        Bidirectional-baseline &  71.6 & 71.6 & 67.2 & 68.1 & 29.7 & 20.0 & 33.5 & 21.1 \\ \hline
        Visible only &   73.7  & 72.9 & - & - & 31.6 & 21.0 & - & - \\ 
         Visible-to-Amodal (w/o SP) & 73.9 & 72.9 & 69.4 & 68.9 & 31.5 & 21.1 & 33.7 & 21.1 \\
          \bottomrule
    \end{tabular}
    \vspace{-1em}
    }
    \label{abla:vis2amodal}
\end{table}

\subsubsection{\textbf{Effectiveness of Visible-to-Amodal Modeling}}
\begin{table}[!t]
\vspace{-0.75em}
    \caption{Ablation study on \textit{IoU performance }with \textit{various configurations of the Cat-SP Retrieve}r, namely using augmented data for training (Aug.), using object category (Cat.)}
    \vspace{-0.5em}
    \centering
    \setlength{\tabcolsep}{10pt}
    \renewcommand{\arraystretch}{.9}
    \resizebox{.45\textwidth}{!}{
    \begin{tabular}{cc|cccc}
    \toprule
         {Cat.} &  {Aug.}  &  {\shortstack{KINS}} &  {\shortstack{D2SA }} &  {\shortstack{COCOA-cls }} &  {\shortstack{COCOA}} \\ \midrule
        \xmark & \xmark & 93.34 & 93.42 & 85.24 & 85.95 \\ 
        \xmark & \checkmark & 94.08 & 94.51 & \textbf{86.25} & \textbf{86.62} \\ 
        \checkmark & \xmark & 94.01 & 94.32 & 85.17 & - \\ 
        \checkmark & \checkmark & \textbf{94.14} & \textbf{95.31} & 86.12 & -
        \\ \bottomrule
    \end{tabular}
    \vspace{-3cm}
    }
    \label{abla:shape_prior}
\end{table}

In \cref{abla:vis2amodal}, we assess the efficacy of visible-to-amodal transition compared to bidirectional learning baseline with ResNet-50 backbone. 
The baseline is implemented as in \cref{fig:ablation_baseline}, which shares the same design with Vis-Occ Mask Head but includes the integration integration of the amodal embedding and amodal mask prediction to enable bidrectional relationship.
The result of Vis-Occ Mask Head is obtained by training it separately from ShapeFormer, showing that dropping the amodal-to-visible relation in the baseline results in better visible segmentation. Moreover, the performance of  ShapeFormer-w/o SP (we remove the use of shape prior for fair comparison with the baseline) illustrates that our design of ShapeFormer does not affect the visible result produced by the Vis-Occ Mask Head, hence results in the enhanced visible-to-amodal feature and final amodal segmentation result.

\subsubsection{\textbf{Effectiveness of Cat-SP Retriever}} 
In \cref{abla:shape_prior}, we examine the benefits of category-specific input for retrieving the shape prior and generating augmented data to enhance training generalization. Our findings indicate that using augmented data during training improves IoU scores across all datasets: 0.74 on KINS, 1.15 on D2SA, 1.01 on COCOA-cls, and 0.67 on COCOA. Regarding the use of category information, we observe a performance improvement when incorporating category information for KINS (0.67 IoU) and D2SA (0.9 IoU). Using category information does not result in performance gains for COCOA-cls. This could be attributed to the variation of shapes within same category in COCOA-cls. In the case of COCOA, where we lack category annotations, we denote the corresponding values as (--).
In the final row of the table, we incorporate using both object category and augmented data into the training process, which yields the best performance on KINS and D2SA, and the second-best performance on COCOA-cls.
\begin{table}[!t]
    \caption{Impact of our \textit{shape-prior masked attention} in amodal transformer decoder $\mathcal{D}_a$.}
    \vspace{-.5em}
    \centering
    \setlength{\tabcolsep}{8pt}
    \renewcommand{\arraystretch}{.75}
    \resizebox{0.5\textwidth}{!}{
    \begin{tabular}{c|c|cccc}
    \toprule
        \multirow{2}{*}{\textbf{Datasets}} & \multirow{2}{*}{\shortstack{\textbf{Shape-prior}\\ \textbf{masked attention}}}& \multirow{2}{*}{$AP\uparrow$} & \multirow{2}{*}{$AP_{50}\uparrow$} & \multirow{2}{*}{$AP_{75}\uparrow$} & \multirow{2}{*}{$AR\uparrow$}\\
        & & & & & \\ \midrule
        \multirow{2}{*}{{KINS}} & \xmark & $33.72$ & $57.80$ & $34.74$ & $21.10$ \\ 
        ~ & \checkmark & $\textbf{34.05}$ & $\textbf{58.61}$ & $\textbf{35.74}$ & $\textbf{22.04}$ \\ \midrule
        \multirow{2}{*}{{D2SA}} & \xmark & $69.44$ & $84.25$ & $74.87$ & $68.92$ \\ 
        ~ & \checkmark & $\textbf{71.03}$ & $\textbf{86.05}$ & $\textbf{76.13}$ & $\textbf{69.31}$ \\ \midrule
        \multirow{2}{*}{{COCOA}} &
         \xmark     & 34.92 & 62.21 & 35.47 & 9.60 \\
         & \checkmark & \textbf{35.71} & \textbf{62.71} & \textbf{36.64} & \textbf{9.90}  \\ \midrule
        \multirow{2}{*}{{\shortstack{COCOA\\-cls}}} & \xmark & $35.78$ & $\textbf{59.25}$ & $36.79$ & $37.05$ \\ 
        ~ & \checkmark & $\textbf{35.83}$ & $58.82$ & $\textbf{38.85}$ & $\textbf{37.13}$ \\ \bottomrule
    \end{tabular}
    }
    \label{abla:masked_attention}
\end{table}
\begin{figure}[!t]
    \centering
    \includegraphics[width=.40\textwidth]{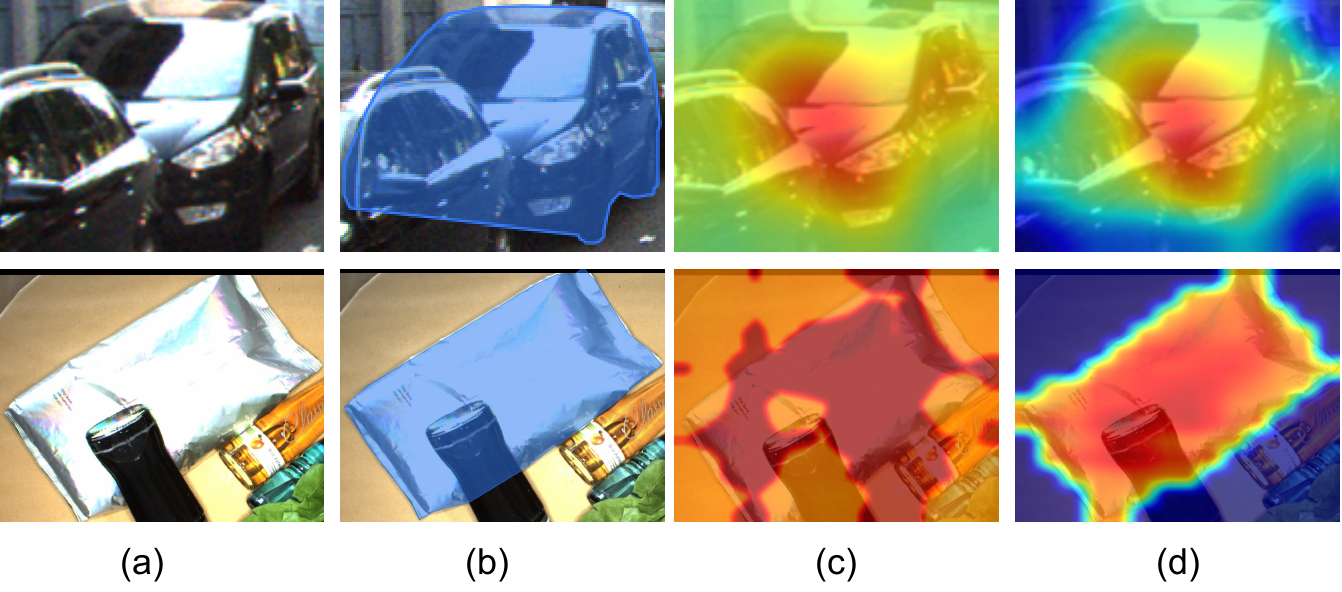}
    \vspace{-0.5em}
    \caption{A visual comparison between using cross attention \cite{vaswani2017attention} and \textbf{our shape-prior masked attention}. From left to right, (a) RoI image, (b) Amodal ground truth mask, (c) Cross attention's attention map, (d) Our shape-prior masked attention's attention map.}
    \vspace{-1.5em}
    \label{fig:attention_viz}
\end{figure}
\subsubsection{\textbf{Effectiveness of shape-prior masked attention}} \cref{abla:masked_attention} showcases the impact of shape-prior masked attention in the Amodal Transformer Decoder $\mathcal{D}_a$. Herein, we evaluate the amodal segmentation performance utilizing the ResNet-50 backbone, where we compare two scenarios: one without shape-prior masked attention (marked as \xmark), and the other with shape-prior masked attention (marked as \checkmark). The results demonstrate that incorporating shape-prior masked attention yields consistent improvements across multiple datasets. These findings highlight the importance of shape-prior masked attention and prior knowledge in enhancing the performance of the Amodal Transformer Decoder $\mathcal{D}_a$ for amodal segmentation. \cref{fig:attention_viz} visualizes the shape-prior masked attention of the Amodal Transformer Decoder on RoIs. 
The attention maps are well-constrained to the object shape owing to the shape-prior masked attention.
Moreover, we can see that the decoder typically attends to the visible parts of objects that are similar to the occluded regions when predicting the amodal mask.

\section{Conclusion}
In conclusion, our proposed ShapeFormer introduces a novel approach to Amodal Instance Segmentation (AIS) by prioritizing the visible-to-amodal transition over the traditional bidirectional method. We address the issue of compromised visible features and present a structured architecture that connects visible and amodal components through shape prior modeling. The transformer-based framework of ShapeFormer leverages advancements in AIS, incorporating a category-specific vector quantized autoencoder for shape prior knowledge. By first predicting visible segmentation while acknowledging occluded objects, and subsequently utilizing shape priors during amodal mask prediction, our model outperforms previous SOTA on AIS across KINS, COCOA, D2SA, COCO-cls datasets. We hope our work sheds light on future
research in AIS aiming to further 
expand the amodal understanding domain.  

\textbf{Acknowledgments}: This work is sponsored by the National Science Foundation (NSF) under Award No OIA-1946391 RII Track-1, NSF 2223793 EFRI BRAID, NSF 2119691 AI SUSTEIN, NSF 2236302, and the National Institutes of Health (NIH) 1R01CA277739-01.

\bibliographystyle{abbrv}
\bibliography{ref}
\end{document}